\documentclass[a4paper,fleqn]{cas-dc}

\usepackage[authoryear,longnamesfirst]{natbib}

\usepackage{url,hyperref,lineno,microtype,subcaption}
\usepackage[onehalfspacing]{setspace}

\usepackage{bm}
\usepackage{algorithmicx,algorithm}
\usepackage[noend]{algpseudocode}
\usepackage{multirow}

\def\tsc#1{\csdef{#1}{\textsc{\lowercase{#1}}\xspace}}
\tsc{WGM}
\tsc{QE}

\begin{document}
\let\WriteBookmarks\relax
\def\floatpagepagefraction{1}
\def\textpagefraction{.001}

\shorttitle{Cooperative trajectory planning algorithm}    

\shortauthors{Huang et al.}  

\title [mode = title]{Cooperative trajectory planning algorithm of USV-UAV with hull dynamic constraints}  



%

\author[1,2]{Tao Huang}
\author[1,2]{Zhe Chen}
\author[3]{Wang Gao}
\author[1,2]{Zhenfeng Xue}[orcid=0000-0002-9593-9429]
\ead{zfxue0903@zju.edu.cn}
\cormark[1]
\author[1,2]{Yong Liu}
\ead{yongliu@iipc.zju.edu.cn}
\cormark[1]

\cortext[1]{Corresponding author}

\affiliation[1]{organization={Institute of Cyber-Systems and Control, Zhejiang University},
            addressline={No.38 Zheda Road}, 
            city={Hangzhou},
            postcode={310027}, 
            country={China}}

\affiliation[2]{organization={Research Center for Intelligent Perception and Control, Huzhou Institute of Zhejiang University},
			addressline={No.819 Xisaishan Road},
			city={Huzhou},
			postcode={313098},
			country={China}}
			
\affiliation[3]{organization={Science and Technology on Complex System Control and Intelligent Agent Cooperation Laboratory},
			addressline={},
			city={Beijing},
			postcode={100191},
			country={China}}


\begin{abstract}
	Efficient trajectory generation in complex dynamic environments remains an open problem in the unmanned surface vehicle (USV). The perception of the USV is usually interfered with by the swing of the hull and the ambient weather, making it challenging to plan the optimal USV trajectories.
In this paper, a cooperative trajectory planning algorithm for the coupled USV-UAV system is proposed to ensure that USV can execute a safe and smooth path in the process of autonomous advance in multi-obstacle maps.
Specifically, the unmanned aerial vehicle (UAV) plays the role of a flight sensor, providing real-time global map and obstacle information with a lightweight semantic segmentation network and 3D projection transformation.
And then, an initial obstacle avoidance trajectory is generated by a graph-based search method.
Concerning the unique under-actuated kinematic characteristics of the USV, a numerical optimization method based on hull dynamic constraints is introduced to make the trajectory easier to be tracked for motion control.
Finally, a motion control method based on NMPC with the lowest energy consumption constraint during execution is proposed.
Experimental results verify the effectiveness of the whole system, and the generated trajectory is locally optimal for USV with considerable tracking accuracy.
\end{abstract}




\maketitle

\section{Introduction}
Unmanned surface vehicles (USVs) are a kind of specific ships with the ability of autonomous mission execution, which are widely used in various applications, including marine resource exploration, water resource transportation, patrol and defense in key areas and river regulation~\cite{wang2020roboat,chen2021novel}, and a large number of research progress have been obtained, including environmental perception~\cite{han2019coastal,cheng2021water}, formation control~\cite{yan2021formation,liu2023distributed}, navigation~\cite{page2022usv,zou2020novel}, and so on.
Environmental perception and trajectory generation are the two most important techniques when the USVs are executing in unknown environments.
Especially when the environment contains dynamic obstacles, the USVs are hard to achieve accurate trajectory planning and tracking due to the lack of effective obstacle information.
As a result, the autonomous navigation system may fail.

During the navigation process of USV, the sensing devices, such as radar or camera, are located at a low observation point, which is detrimental for environmental perception because the adjacent obstacles in the front and behind will block each other.
What's more, the input of the sensors often contains noises caused by hull shaking on the water.
This makes precise environmental perception become a difficult problem for USV and then affect the success rate of trajectory generation.
Usually, simultaneous localization and mapping (SLAM)~\cite{naus2019use} technology is required to construct the global map.
However, this kind of method will bring a huge computational load, and it is intractable to deal with dynamic objects in the water environment.

\begin{figure}[h]
	\centering
	\includegraphics[scale=0.6]{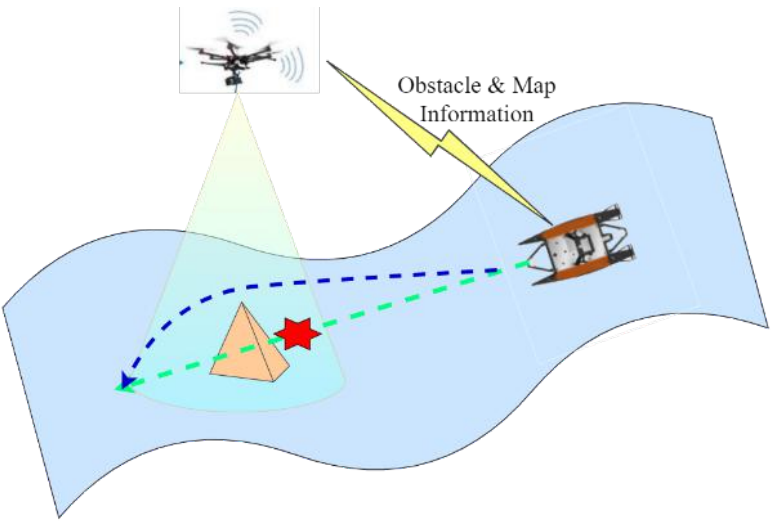}
	\caption{An illustration of the cooperative system of USV-UAV, where the UAV provide wide obstacle and map information to guide the USV to generate obstacle avoidance trajectory.}
	\label{fig:cooperative_system}
\end{figure}

A feasible solution is to design a USV-UAV cooperative system to tackle the above problems, where the unmanned aerial vehicle (UAV) plays the role as a flying sensor.
As shown in Fig.~\ref{fig:cooperative_system}, the USV has long cruise capability, but its perception is disturbed and limited by the circumstance, hence the UAV flies over the USV, providing more stable and comprehensive information.
Semantic segmentation~\cite{chen2017deeplab,yao2021shorelinenet} and 3D projection are used in this paper to transfer obstacle information in the field of vision of UAV to the coordinate system of USV.
Semantic segmentation extracts pixel information of environmental obstacles, and camera projection model helps to transfer pixel information to 3D information.
By doing this, global map information around the USV can be obtained efficiently and in real time, implying the USV-UAV cooperative systems can improve the perception ability of USV effectively, allowing USV to perform tasks in more complex water circumstance.

An initial obstacle avoidance trajectory is firstly generated by a graph-based search method~\cite{niu2019voronoi}.
However, such a method was originally designed for path searching on vast geographical scenarios, which does't consider USV's dynamic characteristics obviously.
On the other hand, USV is famous for its under-actuated motion characteristics~\cite{fossen2021handbook}, which makes it hard to be controlled well even an optimal trajectory is planned.
In this paper, we design a numerical optimization method to optimize the trajectory.
Specifically, we take the hull dynamic constraints into account when modelling the optimization problem.
As a result, the generated trajectory not only allows the obstacle avoidance rule, but also fits the motion characteristics of USV.
This makes the generated trajectory easier to be tracked under the same control conditions.

Finally, a control method with the lowest energy consumption for execution task is designed under a new numerical optimization problem.
It ensures that the power consumption is the optimal when the USV is actuated to track the given optimal trajectory, which is a very useful technique in real-world applications.
The performance of trajectory generation and tracking is comprehensively compared and analysed in the simulation environments, and it verifies the effectiveness of proposed novel framework.

In summary, the contributions of this paper are listed as follows.

\begin{itemize}
	\item A novel USV-UAV cooperative system is proposed, where the UAV acts as a flying sensor to provide global map information around the USV by semantic segmentation and 3D projection, providing more comprehensive and effective perception results for navigation planning.
	
	\item A numerical optimization problem is formulated during the trajectory generation process. It considers the hull under-actuated dynamic constraints and UAV's perception, which can generate a fuel-saving trajectory in real-time optimization.
	
	\item The lowest energy consumption control law is proposed to track the generated trajectory efficiently and accurately, and extensive experiments are conducted to verify the effectiveness of the USV-UAV cooperative system.
\end{itemize}

\section{Related works}

\subsection{Trajectory planning for USV}

Trajectory planning aims to automatically generate an obstacle avoidance trajectory for USV when the local or global map is given.
Among existing methods, the mainstream trajectory planning methods are mainly divided into two categories, \textit{i.e.}, path search and trajectory generation.

For the path search methods, there exist two research directions, including graph search and random sampling.
Typical graph search methods include the A*~\cite{rana2011star} and Dijkstra~\cite{wang2011application} algorithm as well as their derivatives~\cite{zheng2019agv}.
These methods mainly discretize the known map into interconnected grids, and find the shortest path according to the heuristic parameters.
The disadvantage of this kind of method is that the dimension of search in the large map is exploding, and the calculation time shows a rapid upward trend.
Among random sampling methods, typical ones including RRT~\cite{kuffner2000rrt} and its derivatives~\cite{guo2022global}, dynamically find feasible paths by random sampling feasible points in the map and constructing exploratory random trees.
The method can show better performance in large maps, but its shortcomings are also very obvious. 
It is easy to be guided to local optimization, and it is difficult to generate feasible paths in narrow areas when the computing resources of the system are limited.
The common problem of the above methods is that the generated path curvature is discontinuous, and trajectory smoothing is needed afterwards.

For the trajectory generation methods, curve interpolation methods such as B-spline~\cite{zhang2021real} are commonly used to smooth the trajectory.
The smoothness of trajectory and motion state is guaranteed by the continuity theorem of higher-order derivative of curve.
Meanwhile, numerical optimization methods are also widely used, such as Minimum Snap~\cite{mellinger2011minimum} and near-optimal control~\cite{zhang2021learning}.

There are also some methods can combine path search with trajectory generation, such as domain reduction‑based RRT*~\cite{wen2020online} and Hybrid A*~\cite{sedighi2019guided}.
In this paper, the proposed method belongs to the numerical optimization method, and it adds the dynamic constraints and kinematic constraints of unmanned craft in the trajectory generation part, so that the generated trajectory is more in line with the dynamic characteristics of the hull.

\subsection{Cooperative system of USV-UAV}

With the rapid development of automation technology and artificial intelligence technology, unmanned aerial vehicle (UAV) technology has made great progress in these years.
Compared with USV, the advantage of UAV is that it has a broader field of vision and faster movement speed, and can provide more comprehensive and effective data information for USV.
In addition, UAV has the advantage of flying height, and its communication ability is less affected by the environment. 
It can be used to provide communication relay services for multiple USVs located in different positions. 
Due to the strong complementarity between USV and UAV in perception, communication, operation time and other aspects, researchers have made a lot of research on the coordination of UAV serving USV, and successfully verified that this method can effectively solve the above-mentioned problem of self-awareness of USV.

\cite{ozkan2019rescue} focused on the search and rescue of USVs in flood scenes and proposed a collaborative mode of manipulating a UAV to establish the global map first, providing complete map information and target location for subsequent USV planning.
\cite{xue2021distributed} proposed a cooperative formation control algorithm for a single USV and multiple UAVs. The method is based on the leader-follower distributed consensus model, and the position and orientation of each boat are determined by the RGB image color space features acquired by the UAV camera.
\cite{wu2019coordinated} considered the strong search capability of the UAV in the air, combined with the actual target strike capability of the USV, and proposed a two-stage cooperative path planning algorithm on the water and underwater based on the particle swarm optimization algorithm. 
\cite{liu2021uav} proposed an effective game incentive mechanism for the task assignment problem in the cooperative operation of USVs and UAVs, which reduced the task cost and improved the task efficiency. 
 \cite{li2022robust} proposed the LVS-LVA framework to be applied to the cooperative motion control of USV-UAV.


Although most of these methods are  cooperative ways to provide environmental data by UAV and provide perceptual information for the navigation task of USV.
With the development of computer vision technology, the accuracy and robustness of the perception algorithm they use need to be improved. 
In addition, they did not consider the trajectory of USV and its tracking control link, and the proposed collaborative framework can not be fully applicable to the autonomous navigation task of USV.

\section{Cooperative Trajectory generation}

In the cooperative system of USV-UAV, the USV has stable environmental self-supporting ability, and the UAV is flexible and environmentally adaptable.
In the process of autonomous navigation of USV, relying on the wide field of vision and strong environmental perception provided by UAV, it can generate a more reasonable trajectory and skillfully avoid various kinds of obstacles.

\subsection{Environmental perception and 3D projection}

\begin{figure}[h]
	\centering
	\includegraphics[scale=0.4]{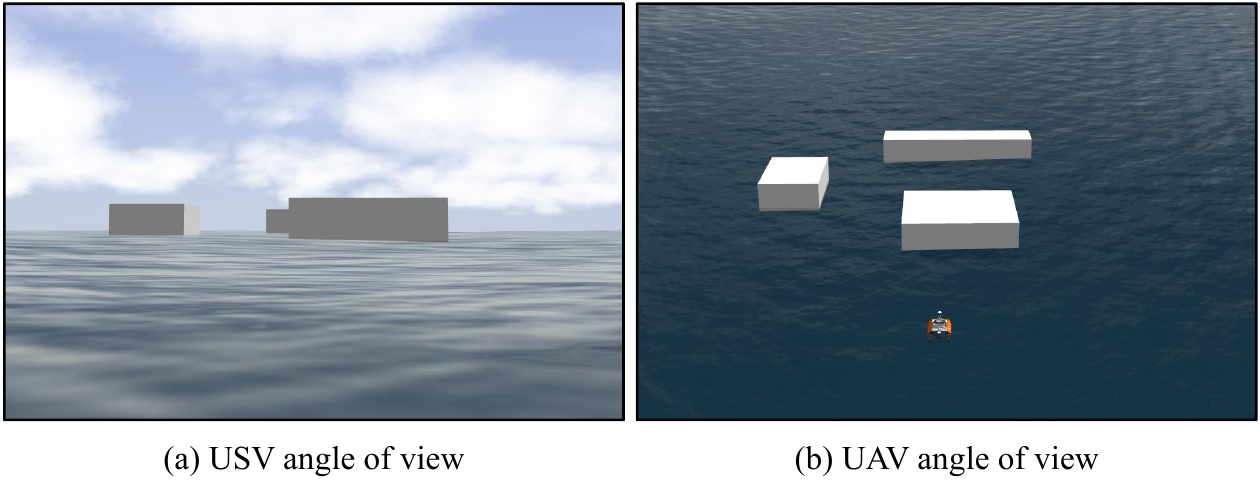}
	\caption{Perspective difference between USV and UAV. The USV observes the environment from a horizontal perspective, leading to serious visual occlusion. Differently, the UAV performs environmental perception from a top-down perspective, and enables to get more accurate map information.}
	\label{fig:different_viewpoint}
\end{figure}

Environmental perception is of vital significance when the USV is executing in an unknown water area.
Different observation angles have a great influence on the observation results.
As shown in Fig.~\ref{fig:different_viewpoint}, the USV and UAV have different angles of view. The USV observes the environment from a horizontal perspective, which may lead to serious visual occlusion.
Differently, the UAV performs environmental perception from a top-down perspective, which enables to get more accurate map information.

Concerning about the accuracy of obstacle recognition and the calculation efficiency, we use semantic segmentation technology~\cite{2015Fully,xue2020ehanet} based on deep learning to extract pixel level obstacle information from the image data obtained by the UAV camera.
For a given image, the position, shape and size of obstacles in the environment can be judged by assigning each pixel with a two-categorical label, where `0' indicates the safety area and `1' denotes the area that the obstacles are located.

\begin{figure*}[h]
	\centering
	\includegraphics[scale=0.48]{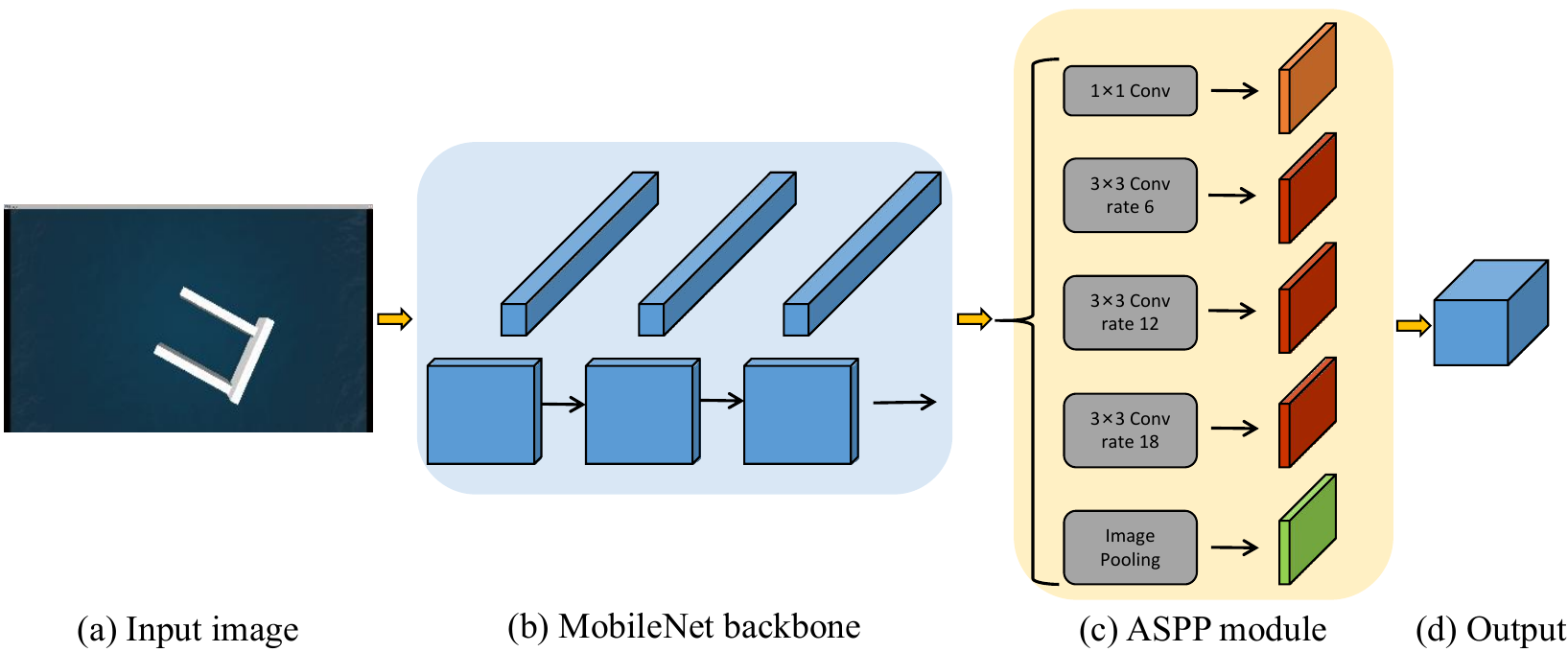}
	\caption{The network architecture of the semantic segmentation algorithm deployed on the UAV.}
	\label{fig:network_architecture}
\end{figure*}

In this paper, we use the DeepLab~\cite{chen2017deeplab} as the semantic segmentation network, and replace the backbone with MobileNet~\cite{qin2018fd}.
On the one hand, it reduces the amount of computation. On the other hand, in the process of feature extraction, with the help of the atrous spatial pyramid pooling (ASPP) module, it can effectively improve the global receptive field and get a better recognition effect. The overall network architecture is illustrated in Fig.~\ref{fig:network_architecture}.

After obtaining the pixel coordinates of obstacles in the image, it needs to convert the obstacle coordinate information into a unified global coordinate.
Let's define the coordinate system of UAV as $U$, camera coordinate as $C$, and the global coordinate as $G$.
Then the transformation from $U$ to $C$ can be represented by $T_{UC}=[R|T]\in R^{4 \times 4}$, where $R$ is the rotation matrix and $T$ is the translation matrix.
$T_{GU} \cdot T_{UC}$ denotes the transformation matrix from $G$ to $C$.
Assuming that the coordinates of the obstacle point $m$ in the pixel coordinate are $(u,v)$, according to the imaging principle of the pinhole camera model, the relationship between its position in the camera coordinate can be expressed as
\begin{equation}
	\left\{
	\begin{aligned}
		u &= f_x \cdot \frac{x}{z} + c_x \\
		v &= f_y \cdot \frac{y}{z} + c_y,
	\end{aligned}
	\right.
\end{equation}
where $f_x$ and $f_y$ denote the focal length in the $x$ and $y$ direction. $c_x$ and $c_y$ are the positions of the origin of the image plane, which can usually be regarded as the center of the image.
Thus, the relationship between the 3D points in the global coordinate $M=(x,y,z)$ and the pixel coordinate $m=(u,v)$ is denoted by
\begin{equation}
	s \cdot \begin{bmatrix}
	u\\
	v\\
	1\\
    1  
	\end{bmatrix} = \begin{bmatrix}
	f_x & 0 & c_x \\
	0 & f_y & c_y \\
	0 & 0 & 1     \\
    0 & 0 & s 
	\end{bmatrix} \cdot T_{GU} \cdot T_{UC} \cdot \begin{bmatrix}
	x\\
	y\\
	z\\
	1
	\end{bmatrix},
\end{equation}
where $s$ is the scaling factor, which can be regarded as the depth information of each pixel.
In this paper, a binocular camera carried by the UAV is used to obtain the pixel depth s.
Through this way of 3D coordinate projection, the pixel information sensed by the UAV in real-time can be projected into the global coordinate, forming the 3D perception ability of USV to the environment.

\begin{algorithm}[h]
	\caption{Trajectory Search with Hybrid A*} 
	\hspace*{0.02in} {\bf Input:} 
	$x_0$, $x_f$, $map$\\
	\hspace*{0.02in} {\bf Output:}  
	Trajectory $T$
	\begin{algorithmic}[1]
		\State Function Search($x_0$, $x_f$, $map$)
		\State $open$ $\leftarrow$ $\phi$, $close$ $\leftarrow$ $\phi$
		\State $open$.push($x_0$)
		\While{$open$ is not $\phi$}
			\State $x_n$ $\leftarrow$ $open$.pop()
			\State $close$.push($x_n$)
			\If{$x_n$.near($x_f$)}
				\If{reedsheep($x_n$, $x_f$)}
					\State \Return $path$($x_f$)
				\EndIf
			\Else
				\For{$x_{succ}$ $\in$ successor($x_n$)}
					\If{$x_{succ}$.safe() and not exist($x_n$, $close$)}
						\State $g$ $\leftarrow$ $g$($x_n$) + $g$($x_{succ}$, $x_n$)
						\If{not exist($x_{succ}$, $open$) or $g$ $<$ $g$($x_{succ}$)}
							\State pred($x_{succ}$) $\leftarrow$ $x_n$
							\State h($x_{succ}$) $\leftarrow$ Heuristic($x_{succ}$, $x_f$)
							\If{not exist($x_{succ}$, $open$)}
								\State $open$.push($x_{succ}$)
							\Else
								\State $open$.rewrite($x_{succ}$)
							\EndIf
						\EndIf
					\EndIf
				\EndFor
			\EndIf
		\EndWhile
		\State \Return $null$
		
	\end{algorithmic}
\end{algorithm}

\subsection{Initial trajectory generation}

In order to generate an obstacle avoidance trajectory, this paper applies the Hybrid A* algorithm~\cite{2019Guided} to provide an initial path, as shown in Algorithm 1.
Given the initial state of USV $s=(x_0, y_0, \varphi_0)$ and the navigation target state $e=(x_f, y_f, \varphi_f)$, the algorithm first puts the initial state into the open list.
Then it iteratively reads the node with the lowest cost in the open list as the current parent node, and generates the next child node according to the current node state, system motion mode and obstacle map.
Different from A* algorithm, the Hybrid A* algorithm adds the orientation dimension to the coordinate. Therefore, the determination of reaching the target state is that the distance between the coordinates of the node and the target point is less than the threshold of reaching distance, and the collision free Reeds-Shepp curve can be generated through the node state and the target point state.

\section{Trajectory optimization and tracking}

The USV is an under-actuated robot operation system, where the number of control variables of the system is less than the degree of freedom of the system.
In the process of trajectory optimization, if the dynamic constraints of this under-actuated characteristic are added to the optimization process, the optimal trajectory more in line with the characteristics of ship motion can be generated.

\subsection{Trajectory optimization with dynamics}

The motion model of USV is a mathematical model with 6 degrees of freedom when it is complete.
For simplicity, we can ignore the motion of the hull in the heave, roll and pitch directions, and simplify it into a 3 degrees of freedom with surge, roll and yaw, represented by $x$, $y$ and $\varphi$.
The mathematical expression of the hull dynamics can be expressed as
\begin{equation}
	\left\{
	\begin{aligned}
		\dot{\bm\eta} &= \bm{J}(\bm\eta) \bm\nu \\
		\bm{M}\dot{\bm\nu} &= \bm\tau -\bm{C}(\bm\nu)\bm\nu-\bm{D}\bm\nu,
	\end{aligned}
	\right.
\end{equation}
where $\bm{\eta}=(x,y,\varphi)$ $\in R^{3\times1}$ denotes the state variables, and $\bm{\nu}=(u,v,r)$ $\in R^{3\times1}$ denotes the speed variables. $\bm{J}$ $\in R^{3\times3}$ is the transition matrix and $\bm{C}$ $\in R^{3\times3}$ is the Coriolis centripetal force matrix. $\bm{M}$ $\in R^{3\times3}$ is the inertial matrix, and $\bm{D}$ $\in R^{3\times3}$is the damping matrix. $\bm{\tau}=(\tau_u, 0, \tau_r)$ $\in R^{3\times1}$ is the thrust matrix. 
For a catamaran, the thrust matrix can be expressed as
\begin{equation}
	\left\{
	\begin{aligned}
		\tau_u &= T_1 + T_2 \\
		\tau_r &= (T_1 - T_2) \cdot B,
	\end{aligned}
	\right.
\end{equation}
where $T_1$ and $T_2$ are the thrust of two propellers, and $B$ is their distance.
The USV can be viewed as a linear time-invariant (LTI) system. Its state variables $\bm{X}$ and control variable $\bm{\tau}$ can be represented by
\begin{equation}
	\left\{
	\begin{aligned}
		\bm{X} &= [x, y, \varphi, u, v, r]^T \\
		\bm{\tau} &= [\tau_u, 0 , \tau_r]^T.
	\end{aligned}
	\right.
\end{equation}
The system dynamics is as follow
\begin{equation}
	\left\{
	\begin{aligned}
		\dot{x} &= ucos(\varphi) - vsin(\varphi) \\
		\dot{y} &= usin(\varphi) + vcos(\varphi) \\
		\dot{\varphi} &= r \\
		m_{11} & \dot{u} - m_{22}ur + d_{11}u = \tau_u \\
		m_{22} & \dot{v} - m_{11}ur + d_{22}v = 0 \\
		m_{33} & \dot{r} +(m_{22}-m_{11})uv+d_{33}r=\tau_r.
	\end{aligned}
	\right.
\end{equation}

On the basis of Hybrid A*, the global trajectory is optimized twice with the following constraints being established, including position, velocity, angular velocity, control input as well as the waypoint state constraints.
The reference  waypoint state is the suboptimal trajectory obtained by considering the vehicle model, which can only provide the simulated optimal information of obstacle avoidance, heading speed and other controls. In this paper, we choose to consider the state vector error in the optimization objective function as a soft constraint.
The final optimization objective can be represented as
\begin{equation}
\begin{split}
	min \quad \frac{1}{2} \{ \sum_{i=0}^N [(\bm{X}_i - \bm{X}_i^{ref})^T \bm{W}_x(\bm{X}_i - \bm{X}_i^{ref}) + \bm{\tau}_i^T{\bm{W}_\tau}\bm{\tau}_i] \\
	+ \sum_{i=1}^N (\bm{\tau}_i - \bm{\tau}_{i-1})^T{\bm{W}_u}(\bm{\tau}_i - \bm{\tau}_{i-1}) \},
\end{split}
\end{equation}
where $\bm{X}_i^{ref}$ denotes the reference state variables generated by Hybrid A*, and $\bm{W}_x$ $= diag\{50,50,20,15,15,15\}$, $\bm{W}_\tau$ $= diag\{5,0,5\}$, $\bm{W}_u$ $= diag\{3,0,3\}$ represent the positive definite cost weight matrix respectively. Moreover, to ensure the trajectory's adequate accuracy, this paper chooses 0.05s as the sampling period.

We adopt the methods of minimizing the control quantity and minimizing the continuous control difference to ensure that the global trajectory generated by optimization can take into account the trajectory index factors such as the smoothing of control quantity and the minimization of energy consumption at the same time. The overall algorithm flow is shown in algorithm 2.

\begin{algorithm}[h]~\label{al:hybrid_a_star}
	\caption{Global Trajectory Optimization} 
	\hspace*{0.02in} {\bf Input:} 
	${X}_0$, ${X}_f$, $path$\\
	\hspace*{0.02in} {\bf Output:}  
	${X}$
	\begin{algorithmic}[1]
		\State Function OptiTraj($X_0$, $X_f$, $path$)
		\For{$i$ = 0 to N}
			\If{$i$ == 1}
				\State $X(i)$ = $X_0$
			\ElsIf{$i$ == N}
				\State $X(i)$ = $X_f$
			\Else
				\State $X(i)$.$x$ = $path_i$.$x$
				\State $X(i)$.$y$ = $path_i$.$y$
				\State $X(i)$.$\varphi$ = $path_i$.$\varphi$
			\EndIf
		\EndFor
		\State Set constraints $C$
		\State Set Objective Function $J$
		\State Optimize($J$, $path$, $C$, $X$)
		\State \Return $X$
	\end{algorithmic}
\end{algorithm}

\subsection{Tracking control with NMPC}

Nonlinear model predictive control (NMPC)~\cite{magni2009nonlinear} is famous for its ability to improve local tracking precision.
It performs periodic real-time optimization according to the prediction time window to achieve the purpose of iterative control to reduce tracking error.
Through the numerical optimization algorithm proposed above, the global trajectory based on the kinematic and dynamic constraints of USV can be obtained, in which the reference control quantity can be obtained. Therefore, the trajectory optimization is considered to use the error index of control quantity as the optimization target.
Set the current time to be $t_j$, the prediction time window to be $W_n$, the optimization problem as NMPC can be formulated as
\begin{equation}
\begin{split}
	\begin{aligned}
	min \quad \frac{1}{2} \sum_{i=t_j}^{t_j+W_n} [(\bm{X}_i - \bm{X}_i^{ref})^T \bm{W}_{mpcx}(\bm{X}_i - \bm{X}_i^{ref}) \\
	+ (\bm{\tau}_i - \bm{\tau}_{i}^{ref})^T{\bm{W}_{mpc\tau}}(\bm{\tau}_i - \bm{\tau}_{i}^{ref}) \\
	+ (\bm{\tau}_i - \bm{\tau}_{i-1})^T{\bm{W}_{mpcu}}(\bm{\tau}_i - \bm{\tau}_{i-1})],
 	\end{aligned}
\end{split}
\end{equation}
where the first term represents the error between the state variable and the reference state variable, which is mainly used to improve the accuracy of state tracking and maintenance in the process of real-time control. The second term represents the error between the control variable and the reference control variable. This item is to meet the index of the lowest energy consumption. Although this problem has been considered in detail in the correspondence of optimization objectives in global trajectory planning, secondary planning in local tracking control can achieve better results. The third term can improve the smoothness of input variables in actual control and meet the needs of practical application control.
$\bm{W}_{mpcx}= diag\left\lbrace10,10,4,2,2,2\right\rbrace$, $\bm{W}_{mpc\tau}= diag\left\lbrace2,0,2\right\rbrace$, $\bm{W}_{mpcu}= diag\left\lbrace4,0,4\right\rbrace$ represent the positive definite cost weight matrix respectively. And considering the control requirements of real-time and stability, we choose $\bm{W}_{n}$ = 30, the sampling period is 0.05s, and the cycle of the NMPC algorithm call is 0.1s.

\section{Experimental analysis}

In this section, we perform simulation experiments using the open source Otter USV simulator~\cite{lenes2019autonomous} within ROS environment.
The Otter simulator is a catamaran with a size of 2.0m long, 1.08m wide and 1.06m high.
With a weight of 65kg assembled, and with the ability to be disassembled into parts weighing less than 20kg, a single operator can launch the Otter from a jetty, lake or riverside, or the beach.
A PX4 drone autopilot is used as the UAV, which is amounted with a monocular camera.
The Otter USV is travelling within a 200$\times$100 square meter area, placed with many blocks as the obstacles.
We set up several different obstacle terrains to test the crossing ability of the USV-UAV cooperative system.

\subsection{Obstacle recognition ability}

\begin{figure*}[h]
	\centering
	\includegraphics[scale=0.55]{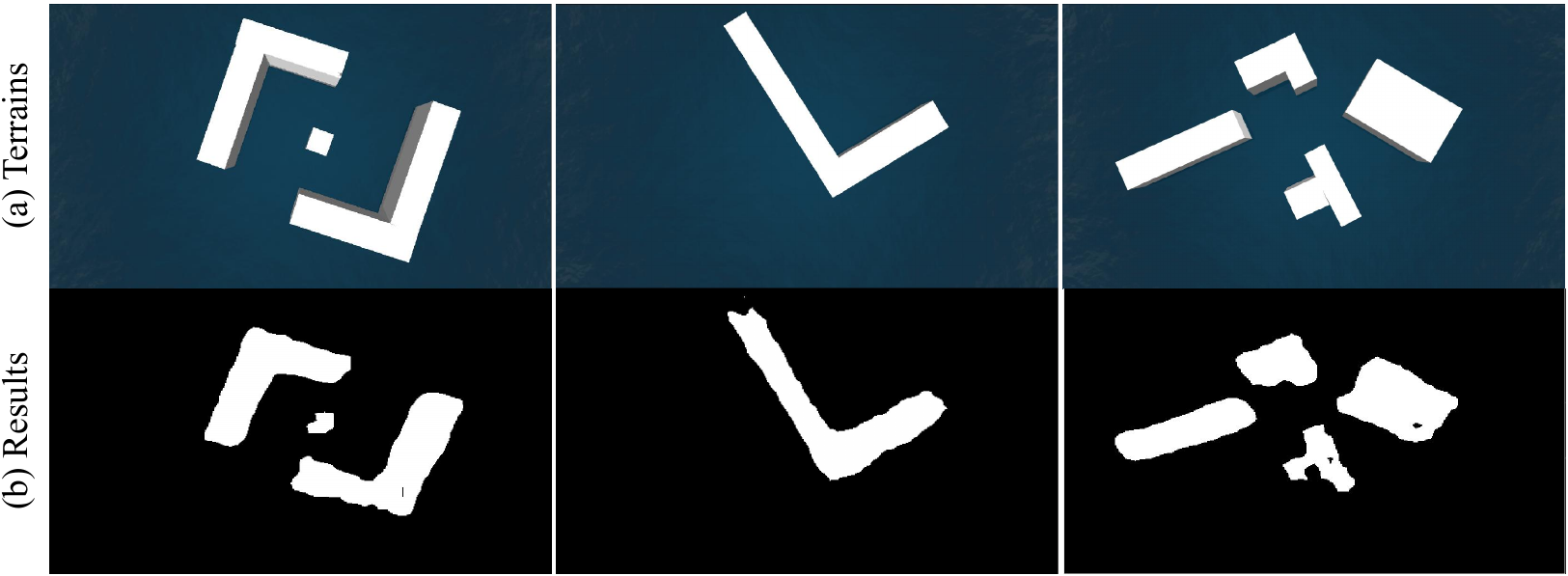}
	\caption{Obstacle recognition results of different terrains.}
	\label{fig:segmentation_result}
\end{figure*}

Firstly, we perform experiments on the ability of obstacle recognition by the USV monocular camera. Semantic segmentation algorithm is used to recognize objects.
Several terrains are randomly placed into the virtual environment.
Some of the segmentation results are shown in Fig.~\ref{fig:segmentation_result}, from which we can see that the proposed light-weight segmentation network can successfully identify obstacles in the environment.
Although there are some empty areas in the middle or the edge of the obstacle, the basic shape of the obstacles has been preserved.
In the post-processing stage, image expansion can be used to increase the safe collision avoidance area and ensure the reliability of navigation.
After that, 3D projection can be performed so that to convert the pixel information into 3D information in global coordinate.

\subsection{Trajectory generation performance}

\begin{figure}[h]
	\centering
	\includegraphics[scale=0.35]{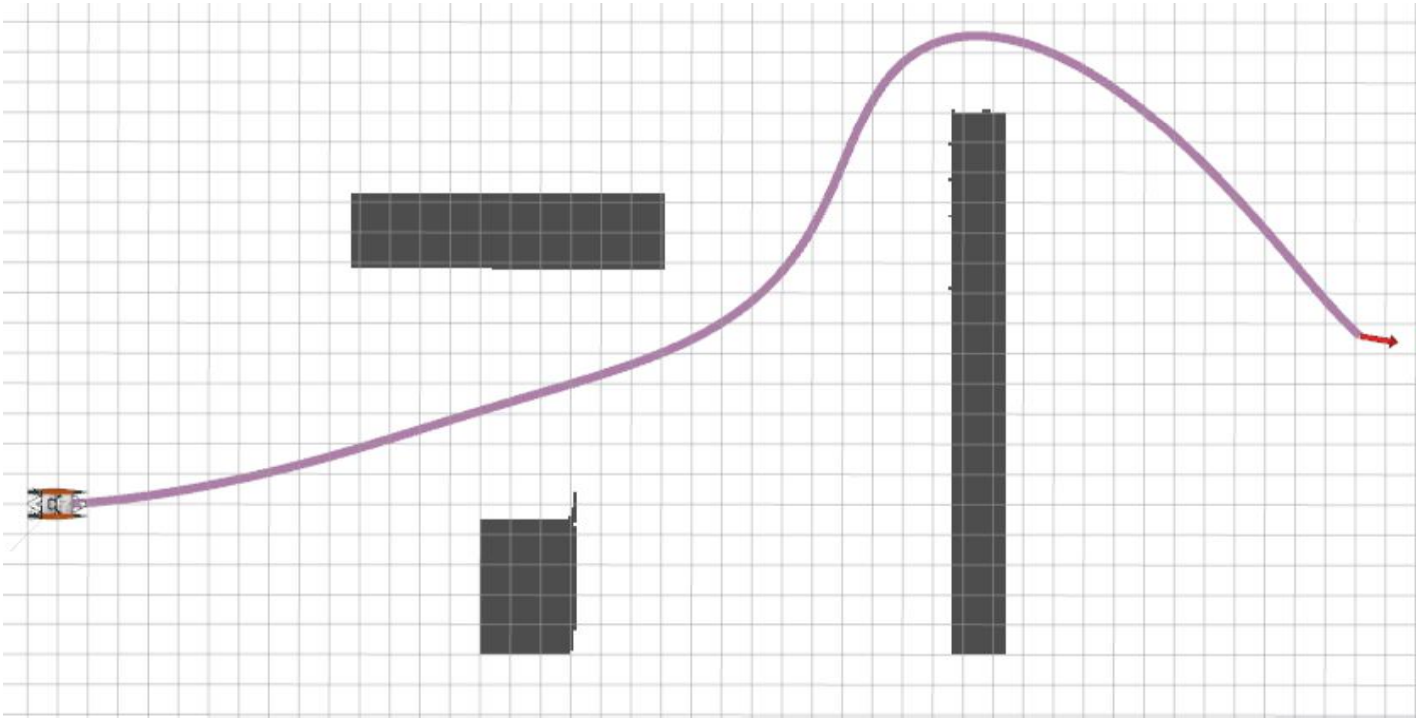}
	\caption{Global trajectory generation performance of USV-UAV cooperative system. The trajectory not only meets collision avoidance condition, but also conform to the kinematic characteristics of the hull.}
	\label{fig:global_trajectory}
\end{figure}

The trajectory generation result is illustrated in Fig.~\ref{fig:global_trajectory}, from which we can see that the generated trajectory not only meets the collision avoidance condition, but also conform to the kinematic characteristics of the hull.
In this paper, the Otter is an under-actuated USV, and it cannot provide direct lateral thrust during its operation.
This requires that the running trajectory of the USV must be smooth enough. Too many bends will bring instability to the motion control of the USV, and then lead to the failure of path trajectory.
The corresponding results can be seen in the subsequent path tracking control experiments.

\begin{figure*}[h]
	\centering
	\includegraphics[scale=0.45]{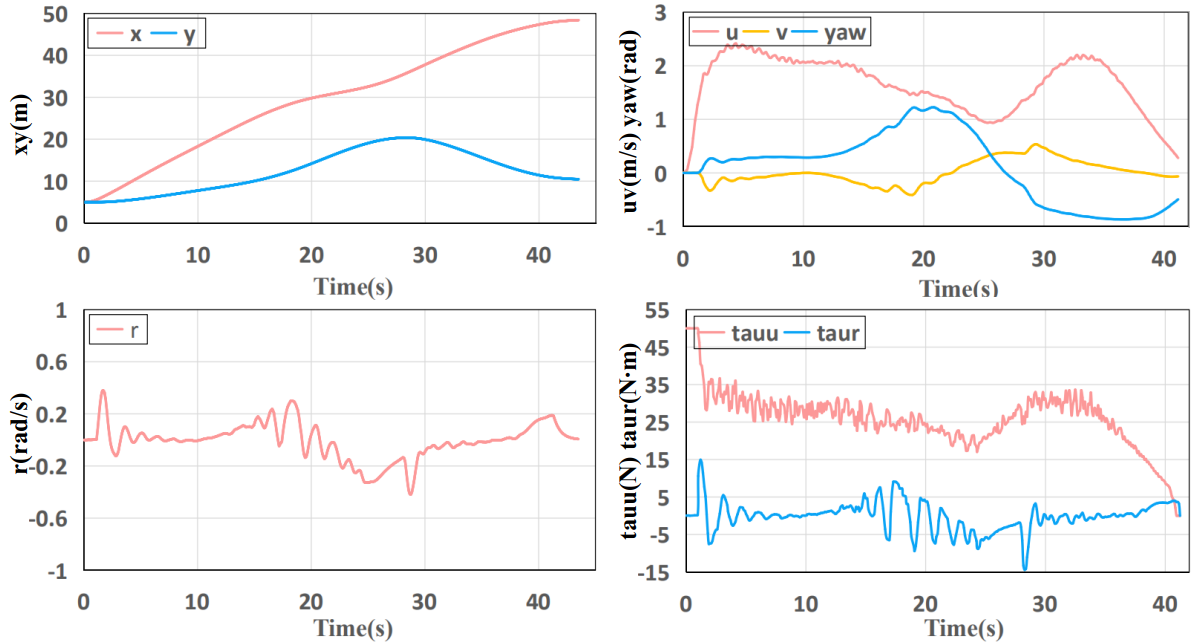}
	\caption{The change trend of the state and control quantity of the USV with time.}
	\label{fig:state_quantity}
\end{figure*}

The change trend of the state and control quantity of the USV with time for the generated trajectory can be found in Fig.~\ref{fig:state_quantity}. Overall, the quantities show a relatively gentle trend, especially for the $x$ and $y$ quantities, which verifies the smoothness of the trajectory.
Higher order quantities such as $u$, $v$ and $yaw$ also present a gentle trend.
This is sufficient to show the effectiveness of the trajectory optimization method.

\begin{figure*}[h]
	\centering
	\includegraphics[scale=0.55]{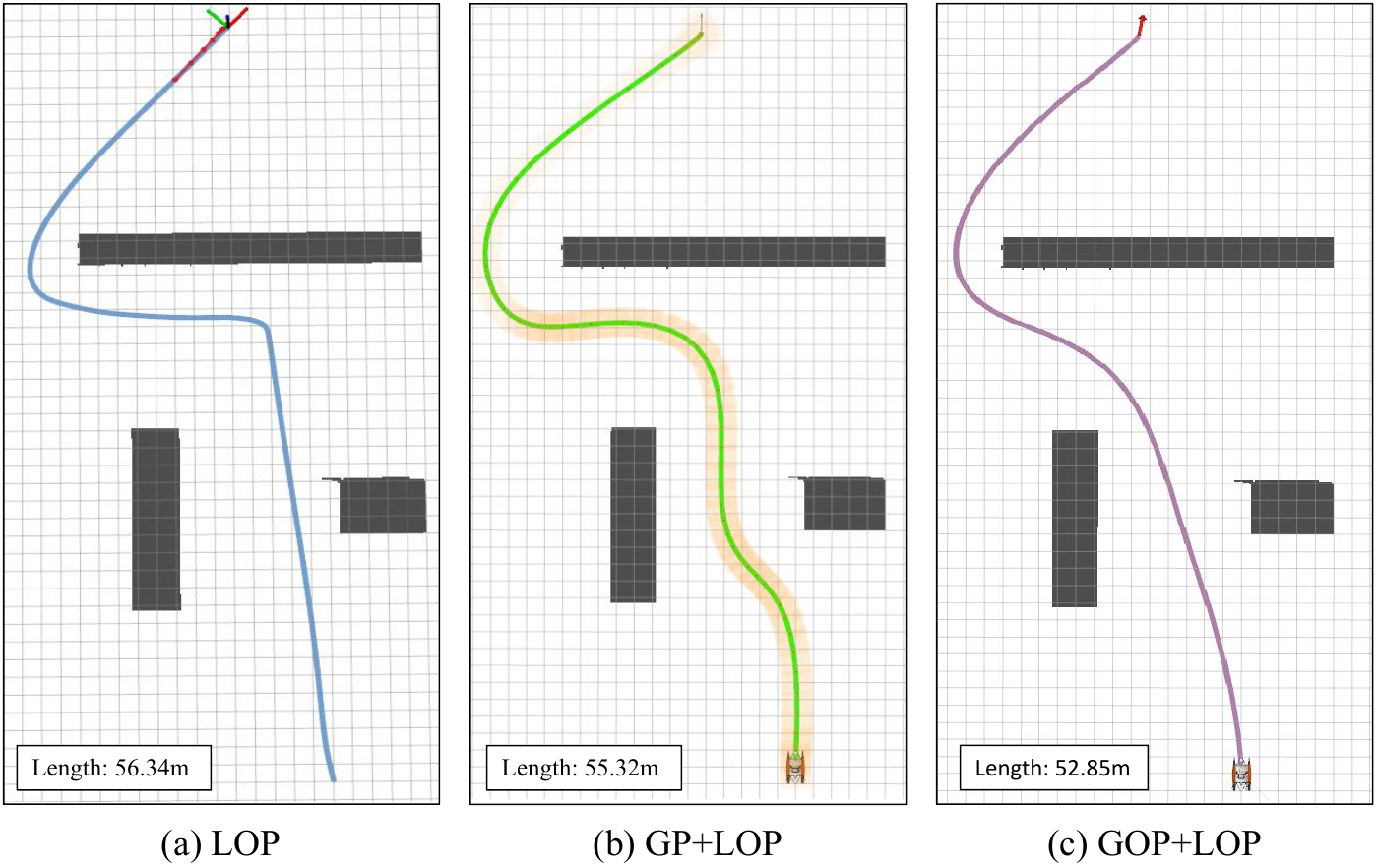}
	\caption{Trajectory generation comparison with different methods. LOP denotes the trajectory generation with local optimization planning, which means the global map provided by UAV is unknown. GP+LOP denotes global planning without trajectory optimization, and GOP+LOP denotes the proposed method.}
	\label{fig:traj_comparison}
\end{figure*}

We also perform ablation study on the proposed method. As shown in Fig.~\ref{fig:traj_comparison}, the LOP and GP+LOP methods are compared. LOP denotes the trajectory generation with local optimization planning, which means the global map provided by UAV is unknown.
Due to the limited perception field of USV, it will take action to perform local trajectory planning unless it is near the obstacle.
GP+LOP denotes global planning without trajectory optimization, which means the global map is known but the trajectory optimization is not performed.
Without the optimization stage, the generated trajectory shows a twisted shape, which is not optimal.
GOP+LOP denotes the proposed method. 
The lower left corner shows the total length of the generated trajectory, and our method obtains the shortest planning path with the best smoothness.

\begin{table*}[h]
	\caption{Quantitative comparison of different trajectory generation methods.}
    \centering
    \begin{tabular}{p{2.4cm}|p{1.8cm}|p{1.8cm}|p{2.0cm}|p{1.8cm}|p{1.8cm}}
    \hline
    \multirow{2}{*}{Method}  & Length   & RMSE     & Max error   & Speed   & Time     \\
          & (m)     & (m)     & (m)      & (m/s)   &  (s) \\
    \hline
    LOP    &  56.34 &  0.120    &  0.3045   &  1.513   & 0.0667 \\
    GP+LOP   & 55.32  &  0.118  & 0.3047  & 1.608  & 0.0697  \\
    GOP+LOP    & \textbf{52.85} & \textbf{0.113}    & \textbf{0.2312} & \textbf{1.675}  & \textbf{0.0506} \\
    \hline
    \end{tabular}
    \label{tab:traj_comparison}
\end{table*}

Here, we also compare the three methods quantitatively in Table~\ref{tab:traj_comparison}. The index such as RMSE, max error, speed and time are evaluated by driving the hull to move.
With the trajectory optimization method, the generated trajectory is more in line with the kinematic characteristics of the hull. As such, the tracking error, execution speed as well as the control time achieves the optimal compared with other methods.

\subsection{Tracking control performance}

\begin{figure*}[h]
	\centering
	\includegraphics[scale=0.5]{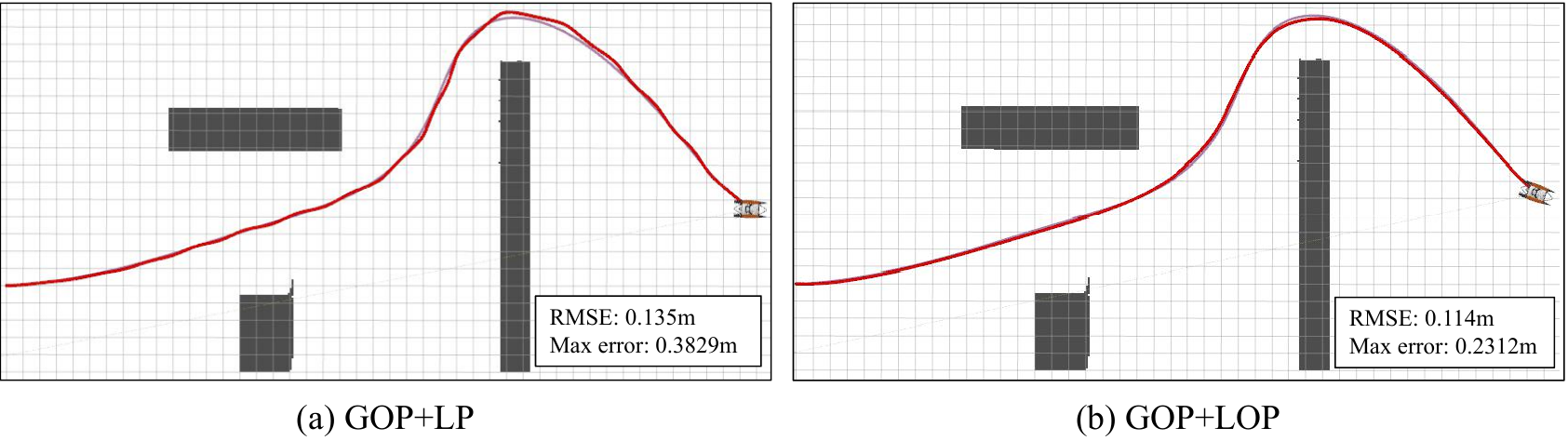}
	\caption{Tracking control performance comparison. GOP+LP denotes the tracking control method without optimization, \textit{i.e.}, the PID control. GOP+LOP denotes the proposed method with NMPC control.}
	\label{fig:tracking_control}
\end{figure*}

To further verify the effectiveness of the proposed NMPC tracking control module, extensive comparative experiments are conducted.
As shown in Fig.~\ref{fig:tracking_control}, GOP+LP denotes the tracking control method without optimization, \textit{i.e.}, the plain PID with adjusted parameters.
The proposed NMPC shows better tracking control performance qualitatively and quantitatively.
There is no prediction time window for GOP+LP, so there will be many small adjustments, resulting in the actual motion trajectory is not smooth.

\begin{figure*}[h]
	\centering
	\includegraphics[scale=0.45]{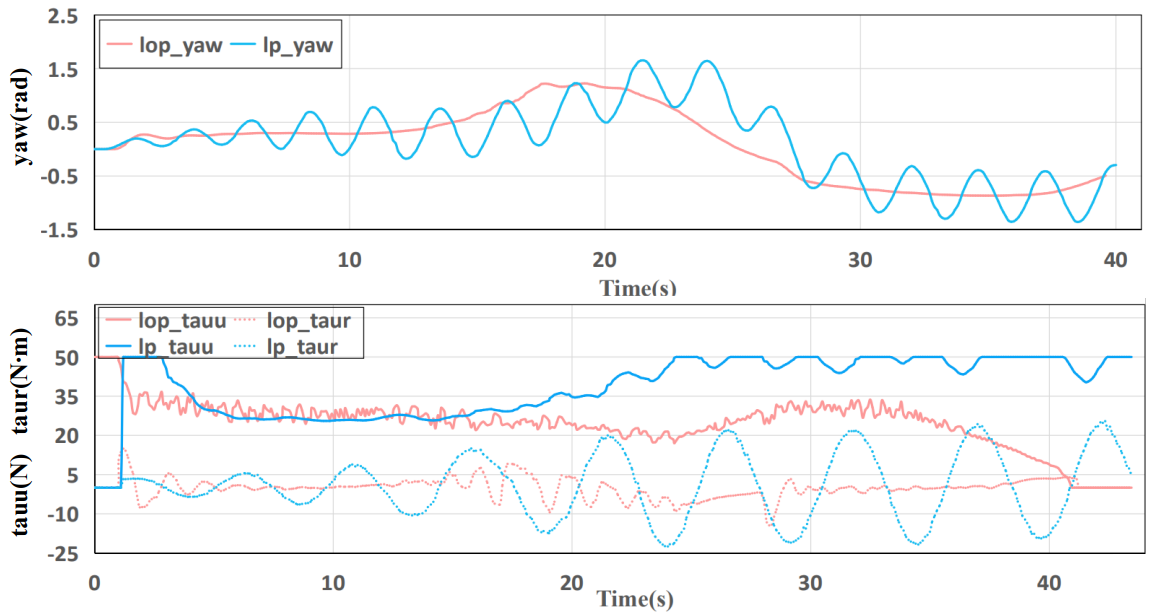}
	\caption{Execution state comparison of motion tracking control.}
	\label{fig:tracking_state}
\end{figure*}

The execution states of different tracking control methods are visualized in Fig.~\ref{fig:tracking_state}, from which the plain PID control shows unstable tracking states. Especially for the control input, the $\tau_r$ shows a divergent trend, which may lead to the input variable to exceed the controllable range and bring adverse effects on the motion control of the USV.

\begin{table}[h]
	\caption{Quantitative comparison of tracking control methods.}
    \centering
    \begin{tabular}{p{1.8cm}|p{1.6cm}|p{1.6cm}|p{1.6cm}}
    \hline
    \multirow{2}{*}{Method}    & RMSE     & Max error   & Speed     \\
          & (m)     & (m)       & (m/s)   \\
    \hline
    GOP+LP    &  0.135  & 0.3829  & 1.327    \\
    GOP+LOP    & \textbf{0.113}    & \textbf{0.2312} & \textbf{1.675}  \\
    \hline
    \end{tabular}
    \label{tab:control_comparison}
\end{table}

Quantitative comparison of tracking control methods can be found in Table~\ref{tab:control_comparison}, from which the proposed method shows better performance compared with GOP+LP (plain PID control).
The proposed method not only achieves smaller tracking control error, but also drives the USV in a quicker speed.
This greatly proves the effectiveness of the combination of motion control and trajectory generation with hull dynamics.

\section{Conclusion}

In this paper, a cooperative trajectory planning algorithm of USV-UAV is proposed to overcome the problem of USV navigation in complex and multi obstacle environment with unknown global map.
The proposed cooperative system is simple yet practical.
In our method, the UAV acts as a flying sensor, providing global map to the USV in real-time with semantic segmentation and 3D projection.
After that, a graph search based method is applied to generate initial obstacle avoidance trajectory.
An optimization method that concerning the kinematic characteristics of the hull is proposed to make the trajectory more in line with the situation.
Finally, a NMPC control method is applied to ensure high precision motion control of USV.
The proposed method has excellent performance and strong practicability in ocean engineering.
In the next step, we will verify the feasibility of the method in the physical experiment and try to study the heterogeneous cooperation scheme of multi USV-UAV systems.

\printcredits

\bibliographystyle{cas-model2-names}
\bibliography{test}

\bio{}
\endbio

\end{document}